\documentclass[conference]{IEEEtran}
\IEEEoverridecommandlockouts
\usepackage{cite}
\usepackage{amsmath,amssymb,amsfonts}
\usepackage{algorithmic}
\usepackage{graphicx}
\usepackage{textcomp}
\usepackage{xcolor}
\def\BibTeX{{\rm B\kern-.05em{\sc i\kern-.025em b}\kern-.08em
    T\kern-.1667em\lower.7ex\hbox{E}\kern-.125emX}}
\usepackage{amsmath}
\usepackage{graphicx}
\usepackage{subfigure}
\usepackage{setspace}
\author{
\IEEEauthorblockN{
Haitong Huang\IEEEauthorrefmark{1}\IEEEauthorrefmark{2},
Cheng Liu\IEEEauthorrefmark{1},
}
\IEEEauthorblockA{
\IEEEauthorrefmark{1}Institute of Computing Technology, Chinese Academy of Sciences, Beijing, P.R.China\\
\IEEEauthorrefmark{2}University of Chinese Academy of Sciences, Beijing, P.R.China\\
}
\IEEEauthorblockA{liucheng@ict.ac.cn}

}
\usepackage{algorithm}
\usepackage{color}

\begin{document}

%%
%% The "title" command has an optional parameter,
%% allowing the author to define a "short title" to be used in page headers.
\title{Deep Learning Accelerator in Loop Reliability Evaluation for Autonomous Driving}

\maketitle

%%
%% The "author" command and its associated commands are used to define
%% the authors and their affiliations.
%% Of note is the shared affiliation of the first two authors, and the
%% "authornote" and "authornotemark" commands
%% used to denote shared contribution to the research.

%%
%% By default, the full list of authors will be used in the page
%% headers. Often, this list is too long, and will overlap
%% other information printed in the page headers. This command allows
%% the author to define a more concise list
%% of authors' names for this purpose.
%\renewcommand{\shortauthors}{Trovato and Tobin, et al.}

%%
%% The abstract is a short summary of the work to be presented in the
%% article.
\begin{abstract}
The reliability of deep learning accelerators (DLAs) used in autonomous driving systems has significant impact on the system safety. However, the DLA reliability is usually evaluated with low-level metrics like mean square errors of the output which remains rather different from the high-level metrics like total distance traveled before failure in autonomous driving. As a result, the high-level reliability metrics evaluated at the post-silicon stage may still lead to DLA design revision and result in expensive reliable DLA design iterations targeting at autonomous driving. To address the problem, we proposed a DLA-in-loop reliability evaluation platform to enable system reliability evaluation at the early DLA design stage. %While accurate pre-silicon DLA simulation with software is time-consuming, we further utilize FPGA to accelerate the simulation to achieve real-time autonomous driving reliability evaluation.

\end{abstract}

%%
%% The code below is generated by the tool at http://dl.acm.org/ccs.cfm.
%% Please copy and paste the code instead of the example below.
%%

%\vspace{-1.5em}

%%
%% Keywords. The author(s) should pick words that accurately describe
%% the work being presented. Separate the keywords with commas.
%\keywords{PIM, Task Scheduling, Heterogeneous}

%%
%% This command processes the author and affiliation and title
%% information and builds the first part of the formatted document.

% For peer review papers, you can put extra information on the cover
% page as needed:
% \ifCLASSOPTIONpeerreview
% \begin{center} \bfseries EDICS Category: 3-BBND \end{center}
% \fi
%
% For peerreview papers, this IEEEtran command inserts a page break and
% creates the second title. It will be ignored for other modes.
%\IEEEpeerreviewmaketitle

\section{Introduction} \label{sec:intro}
Deep learning has become a key technology for autonomous driving. Unlike its deployment in conventional applications, deep learning in autonomous driving has stringent safety requirement. While deep learning models with intensive computation are increasingly deployed on specialized deep learning accelerators (DLA) for both higher performance and energy efficiency, the reliability of the underlying DLAs can dramatically affect the execution of the deep learning models according to the experiments in Section \ref{sec:evaluation}. Hence, reliable DLA design is imperative for functional safety in autonomous driving system. 

Reliability design is not a new concept and there are many fault-tolerant design techniques such as triple modular redundancy (TMR) and ECC that can be used to improve the DLA reliability\cite{2019Design}. Nevertheless, as the influence of faults on the DLAs is closely related with the safety of autonomous driving, the reliability evaluation of the DLA design for autonomous driving system poses new design challenges. First, model level metrics such as model accuracy and intersection of union (IoU) are not always consistent with the safety metrics of the autonomous driving. For instance, when the color of a traffic light is mistakenly recognized due to hardware faults, the consequence of the wrong classifications from red to green and green to red is totally different from the safety point of view, while the influence on the classification accuracy is the same. Unlike the normal DLA design where the deep learning reliability fully depends on the algorithm, hardware faults can corrupt both the DLA functionality and the model reliability. As a result, the DLA reliability evaluation needs to be conducted in an autonomous driving system such that the DLA architecture and the model are evaluated at the same time to ensure the reliability metrics. Second, accurate fault simulation of DLAs can be extremely slow especially for the computation intensive deep learning processing, which essentially hinders the reliability evaluation of the DLAs for autonomous driving. Insufficient reliability evaluation may miss critical bugs and can cause safety problems at the later design iterations, which will incur much more design cost eventually. In summary, a systematical evaluation platform is required for highly reliable DLA design targeting at autonomous driving systems and the evaluation must be fast enough to ensure comprehensive reliability evaluation. 

\iffalse
\begin{itemize}
    \item We propose a hybrid computing architecture for fault-tolerant DLAs. It has a DPPU seated along with the regular computing array to recompute all the processing mapped to faulty PEs in arbitrary locations. DPPU works concurrently with the original computing array and has no accuracy penalty nor performance penalty when the PE fault error rate is less than 3$\%$ . 
    
    \item We also investigate the approaches to degrade both the computing array and the redundant DPPU gracefully to ensure the functionality of the overall computing array under higher fault injection rate.
    
    \item We analyze the reliability of the proposed HCA with comprehensive experiments and the experiments reveal that HCA achieves much higher performance and reliability under various fault injection scenarios with minor chip area penalty when compared to prior redundancy design approaches. 
    
\end{itemize}
\fi
%\input{relatedwork} 
%\input{motivation}
%\input{model}
\section{DLA-in-Loop Reliability Evaluation Platform} \label{sec:overview}
\begin{figure*}
\centering
\begin{minipage}[!htbp]{0.65\linewidth}
\includegraphics[width=4.5in]{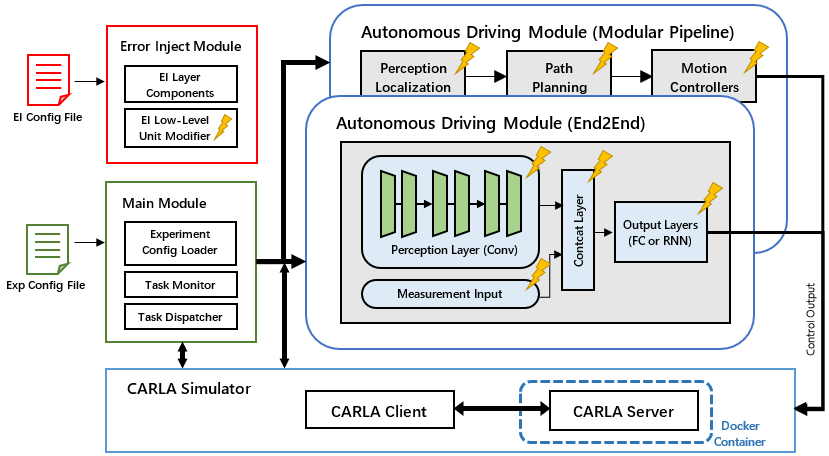}
\caption{Overview of the DLA-in-Loop Reliability Evaluation Platform}
\label{fig:platform}
\end{minipage}%
\begin{minipage}[!htbp]{0.35\linewidth}
\centering
\includegraphics[width=1.6in,height=1.2in]{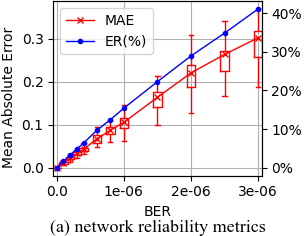}\\[3pt]
\includegraphics[width=1.6in,height=1.2in]{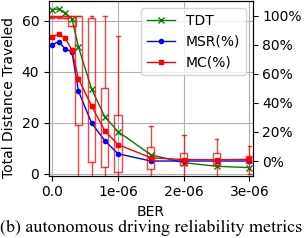}
\caption{Reliability Metrics Comparison}
\label{fig:metric}
\end{minipage}
%\caption{3 figure}
\end{figure*}
To address the above problem, we propose to put the DLA in the simulation loop of an autonomous driving system such that the DLA reliability can be systematically evaluated at the early chip design stage. The proposed reliability evaluation platform as shown in Figure \ref{fig:platform} is designed to be as modular and generic as possible for testing a variety of different autonomous driving systems. It has CARLA\cite{2017CARLA}, an open-source autonomous driving simulator, to provide autonomous driving performance evaluation. It has a master process for reading the driving test configuration, running parallel test tasks according to the system resources, and monitoring the status of the tasks. It also includes many sub processes for CARLA server, autonomous driving agent, experiment data logging, and so on. In addition, it has a simulation error injection module designed to be independent of the autonomous driving agent. In this case, the neural network module in agent can be replaced with our fault evaluation module with only minor modification.

CARLA supports different autonomous driving modules such as the end-to-end autonomous driving and the modular pipeline based autonomous driving. Typically, it has GPUs to conduct the neural network processing used in the autonomous driving modules. To evaluate the pre-silicon DLA reliability in autonomous driving system, we have the deep learning tasks performed on the DLA with fault injection instead, which can be a cycle-accurate simulator. Then we can inspect the behaviors of the autonomous driving directly and evaluate the system level metrics such as total distance travelled before failure with the task monitor. Compared to the neural network model level metrics like classification accuracy, the high-level evaluation reveals the actual influence of the hardware faults in DLAs on the autonomous driving safety, which is more appropriate to guide the reliable DLA design. Nevertheless, the major challenge of the DLA-in-Loop simulation is that the low-level cycle accurate DLA simulation can be extremely time-consuming. To that end, we further have the DLAs implemented on FPGAs. At the same time, the fault injection is also replaced with the hardware implementation on FPGAs such that the entire DLA-in-loop evaluation platform can be conducted at real-time. 

\section{Experiment} \label{sec:evaluation}
In the experiments, we mainly investigate the difference between the reliability metrics of the network and the autonomous driving caused by the hardware faults in DLAs. We utilize an End2End Imitation Learning (IL) model\cite{2019Exploring} as an example for the autonomous driving. The perception module in IL model is ResNet34 and the back end of the IL model is full connection that outputs control signals including steer, throttle, brake to the simulated vehicle. Note that the signals normally range from -1 to 1. We choose Town01 as the test map and a normal traffic density in CARLA. Then we construct 100 driving missions that contain 25 paths and 4 types of weather
as the benchmark. In each mission, the vehicle performs the autonomous driving from a starting point to an end point. Due to the DLA simulation speed limitation, we only inject random single even upset (SEU) faults to the inputs, weights and hidden states of the neural network models rather than the detailed DLA simulator for preliminary evaluation. The bit error rate (BER) is defined as the probability that a bit in inputs, weights, and hidden states flips.

We have both network level metrics and autonomous driving metrics to evaluate the influence of the DLA faults. They are defined as follows.
Two network level metrics are
\begin{itemize}
    \item \emph{Error Rate (ER)}: the probability that the network output differs from that of the golden run. The difference threshold is set to be 0.01 and smaller difference is ignored.
    \item \emph{Mean Absolute Error (MAE)}: The average absolute error of the results with golden run.
\end{itemize}
The autonomous driving metrics are 
\begin{itemize}
    \item \emph{Mission success rate (MSR)}: The probability that the vehicle completes the mission without collision.
    \item \emph{Mission completion (MC)}: The percentage of the travelled distance over the total distance in a mission.
    \item \emph{Total distance traveled (TDT)}: the distance traveled by the vehicle before the mission is completed or failed.
\end{itemize}

The reliability evaluation with low-level network metrics and high-level autonomous driving metrics are illustrated in Fig 2. It can be observed that network level reliability metrics change gracefully with the rising BER while driving metrics exhibit much more significant change. In addition, the network level reliability metric has little variation under lower BER in the 100 driving missions. In contrast, the high-level reliability metrics show very large scale variations. It indicates that random faults can still cause non-trivial mission failure even under very low BER, which also demonstrates the great difference between these reliability metrics and the high-level reliability metrics are vital to the reliable DLA design. %We found that the automatic driving system is robust, with 2\% of the control data affected at a low BER (1e-6), and the automatic driving system still maintains proper operation. At further increase in BER, the success rate decreases significantly.

%Considering that random flipping of the exponent part of a floating-point number leads to huge changes in the value, while the change of the significand part has less effect on the result, we conduct partial error-injection experiments assuming that the computation of the exponent is protected by some highly reliable mechanism (e.g., TMR). We found that the reliability and security of the system are substantially improved (Table 1). The results show that the reliability of the processor design can be improved at a small cost by protecting only a fraction of the important hardware units. 

We also tried to integrate a cycle-accurate DLA simulator in CARLA for more accurate system level reliability evaluation. Nevertheless, the low-level simulation is more than two or three orders of magnitudes slower than the normal neural network processing on GPUs. As a result, the autonomous driving simulation cannot finish the experiment in more than a week. Therefore, FPGA acceleration is imperative for the pre-silicon DLA reliability evaluation targeting at the safety autonomous driving system.

\bibliographystyle{IEEEtran}
\bibliography{refs}

% trigger a \newpage just before the given reference
% number - used to balance the columns on the last page
% adjust value as needed - may need to be readjusted if
% the document is modified later
%\IEEEtriggeratref{8}
% The "triggered" command can be changed if desired:
%\IEEEtriggercmd{\enlargethispage{-5in}}

% references section

% can use a bibliography generated by BibTeX as a .bbl file
% BibTeX documentation can be easily obtained at:
% http://mirror.ctan.org/biblio/bibtex/contrib/doc/
% The IEEEtran BibTeX style support page is at:
% http://www.michaelshell.org/tex/ieeetran/bibtex/
%\bibliographystyle{IEEEtran}
% argument is your BibTeX string definitions and bibliography database(s)
%\bibliography{IEEEabrv,../bib/paper}
%
% <OR> manually copy in the resultant .bbl file
% set second argument of \begin to the number of references
% (used to reserve space for the reference number labels box)
%\begin{thebibliography}{1}

%\bibitem{IEEEhowto:kopka}
%H.~Kopka and P.~W. Daly, \emph{A Guide to \LaTeX}, 3rd~ed.\hskip 1em plus
%  0.5em minus 0.4em\relax Harlow, England: Addison-Wesley, 1999.

%\end{thebibliography}

% that's all folks
\end{document}